\documentclass[runningheads,a4paper]{llncs}

\usepackage{comment}
\usepackage{amsmath}
\usepackage{amssymb}
\usepackage[linesnumbered,ruled,vlined]{algorithm2e}
\usepackage{float}
\usepackage{xfrac}
\usepackage{tikz}
\usetikzlibrary{positioning}
\usepackage{subfig} 
\usepackage{booktabs}

\usepackage[all]{xy}

\usepackage{graphicx}

\newcommand{\keywords}[1]{\par\addvspace\baselineskip
\noindent\keywordname\enspace\ignorespaces#1}

\begin{document}

\mainmatter  % start of an individual contribution

\title{Human Activity Recognition\\using Recurrent Neural Networks}

\titlerunning{Human Activity Recognition\\using Recurrent Neural Networks}

%the name(s) of the author(s) follow(s) next

\author{Deepika Singh\inst{1}, Erinc Merdivan\inst{1}, Ismini Psychoula\inst{2}, Johannes Kropf\inst{1}, Sten Hanke\inst{1},  Matthieu Geist\inst{3}, \and Andreas Holzinger\inst{4}}
\authorrunning{Deepika Singh et al.}
\tocauthor{Deepika Singh, Erinc Merdivan, Ismini Psychoula, Johannes Kropf,
Sten Hanke, Matthieu Geist, and Andreas Holzinger}
\institute{AIT Austrian Institute of Technology, Austria\\ \email{deepika.singh@ait.ac.at, erinc.merdivan@ait.ac.at}\\
\and School of Computer Science and Informatics, De Montfort University, UK\\
%\email{ismini.psychoula@dmu.ac.uk}
\and CentraleSupelec, France\\
%\email{matthieu.geist@centralesupelec.fr}
\and
Holzinger Group, HCI-KDD, Institute for Medical Informatics/Statistics,
Medical University Graz, Austria} %\texttt{andreas.holzinger@medunigraz.at}}
%\mailsa\smallskip
%\mailsb\smallskip
\toctitle{}

\maketitle
\begin{abstract}
Human activity recognition using smart home sensors is one of the bases of ubiquitous computing in smart environments and a topic undergoing intense research in the field of ambient assisted living. The increasingly large amount of data sets calls for machine learning methods. In this paper, we introduce a deep learning model that learns to classify human activities without using any prior knowledge. For this purpose, a Long Short Term Memory (LSTM) Recurrent Neural Network was applied to three real world smart home datasets. The results of these experiments show that the proposed approach outperforms the existing ones in terms of accuracy and performance.

\keywords{machine learning, deep learning, human activity recognition, sensors, ambient assisted living, LSTM}
\end{abstract}

\section{Introduction}
Human Activity recognition has been an active research area in the last decades due to its applicability in different domains and the increasing need for home automation and convenience services for the elderly \cite{RoeckerEtAl:2011:AAL}. Among them, activity recognition in Smart Homes with the use of simple and ubiquitous sensors, has gained a lot of attention in the field of ambient intelligence and assisted living technologies for enhancing the quality of life of the residents within the home environment~\cite{chen2012sensor}. 

The goal of activity recognition is to identify and detect simple and complex activities in real world settings using sensor data. It is a challenging task, as the data generated from the sensors are sometimes ambiguous with respect to the activity taking place. This causes ambiguity in the interpretation of activities. Sometimes the data obtained can be noisy as well. Noise in the data can be caused by humans or due to error in the network system which fails to give correct sensor readings. Such real-world settings are full of uncertainties and calls for methods to learn from data, to extract knowledge and helps in making decisions. Moreover, the inverse probability allows to infer unknowns and to make predictions \cite{Holzinger:2017:InauguralMAKE}.

Consequently, many different probabilistic, but also non-probabilistic models, have been proposed for human activity recognition. Patterns corresponding to the activities are detected using sensors such as accelerometers, gyroscopes or passive infrared sensors, \textit{etc.}, either using feature extraction on sliding window followed by classification~\cite{roggen2015limited} or with Hidden Markov Modeling (HMM)~\cite{duong2005activity}.

In recent years, there has been a growing interest in deep learning techniques. Deep learning is a general term for neural network methods which are based on learning representations from raw data and contain more than one hidden layer. The network learns many layers of non-linear information processing for feature extraction and transformation. Each successive layer uses the output from the previous layer as input. Deep learning techniques have already outperformed other machine learning algorithms in applications such as computer vision~\cite{lee2009convolutional}, audio~\cite{lee2009unsupervised} and speech recognition~\cite{hinton2012deep}.   

In this paper, we introduce a recurrent neural network model for human activity recognition. The classification of the human activities such as cooking, bathing, and sleeping is performed using the Long Short-Term Memory classifier (LSTM) on publicly available Benchmark datasets~\cite{kasteren2011human}. An evaluation of the results has been performed by comparing with the standardized machine learning algorithms such as Naive Bayes, HMM, Hidden Semi-Markov Model (HSMM) and Conditional Random Fields (CRF).

The paper is organized as follows. Section~\ref{RelatedWork} presents an overview of activity recognition models and related work in machine learning techniques. Section~\ref{LSTMModel} introduces Long Short-Term Memory (LSTM) recurrent neural networks. Section~\ref{experiments} describes the datasets that were used and explains the results in comparison to different well-known algorithms. Finally, Section~\ref{Discussion} discusses the outcomes of the experiments and suggestions for future work.

\section{Related work}
\label{RelatedWork}
In previous research, activity recognition models have been classified into data-driven and knowledge-driven approaches. The data-driven approaches are capable of handling uncertainties and temporal information~\cite{yuen2010data} but require large datasets for training and learning. Unfortunately, the availability of large real world datasets is a major challenge in the field of ambient assisted living. The knowledge-driven techniques are used in predictions and follow a description-based approach to model the relationships between sensor data and activities. These approaches are easy to understand and use but they cannot handle uncertainty and temporal information~\cite{ye2015kcar}.

Various approaches have been explored for activity recognition, among them the majority of the techniques focuses on classification algorithms such as Naive Bayes (NB)~\cite{tapia2004activity}, Decision Trees~\cite{bao2004activity}, HMM~\cite{duong2005activity} , CRF~\cite{van2008accurate}, Nearest Neighbor (NN)~\cite{wu2012classification}, Support Vector Machines (SVM)~\cite{zhu2013context} and different boosting techniques.

A simple probabilistic classifier in machine learning is the Naive Bayes classifier which yields good accuracy with large amounts of sample data but does not model any temporal information. The HMM, HSMM, and CRF are the most popular approaches for including such temporal information. However, these approaches sometimes discard pattern sequences that convey information through the length of intervals between events. This motivates the study of recurrent neural networks (RNN) which promises the recognition of patterns that are defined by temporal distance~\cite{ribbe1997nursing}.

LSTM is a recurrent neural network architecture that is designed to model temporal sequences and learn long-term dependency problems. The network is well suited for language modeling tasks; it has been shown that the network in combination with clustering techniques increases the training and testing time of the model~\cite{sundermeyer2012lstm} and outperforms the large scale acoustic model in speech recognition systems~\cite{sak2014long}.

\section{LSTM Model}
\label{LSTMModel}
\begin{figure}[H]
\centering
  \includegraphics[scale=1.0]{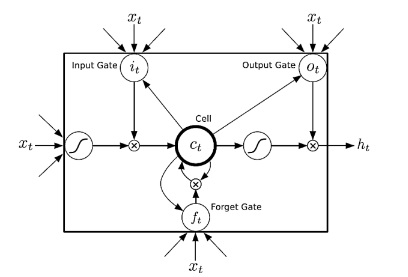}
  \caption{LSTM single cell image~\cite{zhang2015bidirectional}.}
  \label{fig: LSTM}
\end{figure}

LSTM is a recurrent neural network architecture that was proposed in~\cite{hochreiter1997long}. Another version without a forget gate was later proposed in~\cite{gers2000learning} and extended in~\cite{gers2002learning}. LSTM has been developed in order to deal with gradient decay or gradient blow-up problems and can be seen as a deep neural network architecture when unrolled in time. The LSTM layer's main component is a unit called memory block. An LSTM block has three gates which are input, output and forget gates. These gates can be seen as write, read and reset operations for the cells. An LSTM cell state is the key component which carries the information between each LSTM block. Modifications to the cell state are controlled with the three gates described above. An LSTM single cell, as well as how each gate is connected to each other and the cell state itself, can be seen in Figure~\ref{fig: LSTM}.

Each gate and cell state are governed by multiplicative equations that are given by:
\begin{align*}
 i_{t} &= \sigma(W_{xi}x_{t}  +  W_{hi}h_{t-1} + W_{ci}c_{t-1} +  b_{i}),\\
 f_{t} &= \sigma(W_{xf}x_{t}  +  W_{hf}h_{t-1} + W_{cf}c_{t-1} +  b_{f}),\\
 o_{t} &= \sigma(W_{xo}x_{t}  +  W_{ho}h_{t-1} + W_{co}c_{t} +  b_{o}),\\
 c_{t} &= f_{t}c_{t-1} + i_{t}\tanh(W_{xc}x_{t} + W_{hc}h_{t-1} + b_{c}),\\
 h_{t} &= o_{t}\tanh c_{t},
\end{align*}
with $W$ being the weight matrix and $x$ is the input, $\sigma$ being the sigmoid and $\tanh$ is the hyperbolic tangent activation function. The terms $i$, $f$ and $o$ are named after their corresponding gates and $c$ represents the memory cell~\cite{zhang2015bidirectional}.
%images
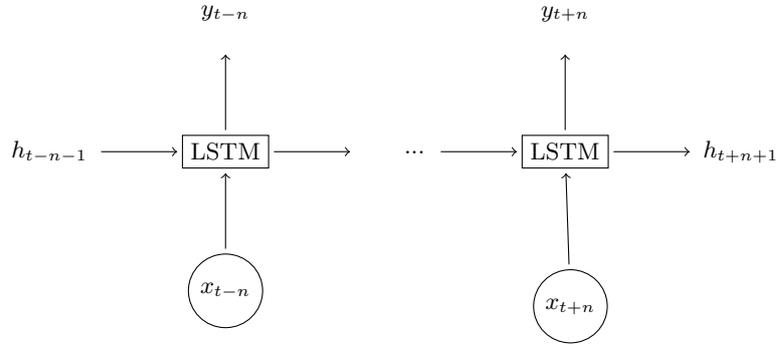
\begin{figure}[tbh]
\centering
\begin{tikzpicture}
\node[draw, outer sep=2, circle] (x_t) {$x_{t-n}$};
\node[draw, align=center, outer sep=2] (unit_t) [above=of x_t] {LSTM};
\node[align=center, outer sep=2] (unit_tprev) [left=of unit_t] {$h_{t-n-1}$};
\node[align=center, outer sep=2] (unit_tnext) [right=of unit_t]{} ;
\node[outer sep=2, circle] (y_t) [above=of unit_t] {$y_{t-n}$};
\node[outer sep=2, circle] (y_tm) [right=0.15cm of unit_tnext] {$...$};

\node[draw, outer sep=2, circle] (x_t1)[below right=2cm of y_tm] {$x_{t+n}$};
\node[draw, align=center, outer sep=2] (unit_t1) [right=of y_tm]{LSTM};
% \node[align=center, outer sep=2] (unit_tprev1) [left=of unit_t1] {$h_{t-1}$};
\node[align=center, outer sep=2] (unit_tnext1) [right=of unit_t1] {$h_{t+n+1}$};
\node[outer sep=2, circle] (y_t1) [above=of unit_t1] {$y_{t+n}$};

\path[->] (x_t) edge (unit_t);
\path[->] (unit_t) edge (y_t);
\path[->] (unit_tprev) edge (unit_t);
\path[->] (unit_t) edge (unit_tnext);

\path[->] (y_tm) edge (unit_t1);
\path[->] (x_t1) edge (unit_t1);
\path[->] (unit_t1) edge (y_t1);
\path[->] (unit_t1) edge (unit_tnext1);
\end{tikzpicture}

\caption{Illustrations of an LSTM network with $x$ being the binary vector for sensor input and $y$ being the activity label prediction of the LSTM network.}
\label{fig: illus}
\end{figure}

By unrolling LSTM single cells in time we construct an LSTM layer where $h_{t}$ is the hidden state and $y_{t}$ is the output at time $t$ as shown in Figure~\ref{fig: illus}.

\section{Experiments}
\label{experiments}

\subsection{Dataset}

Publicly available and annotated sensor datasets have been used to evaluate the performance of the proposed approach~\cite{kasteren2011human}. In this dataset, there are three houses with different settings to collect sensory data. The three different houses were all occupied by a single user named A, B, and C respectively. Each user recorded and annotated their daily activities. Different number of binary sensors were deployed in each house such as passive infrared (PIR) motion detectors to detect motion in a specific area, pressure sensors on couches and beds to identify the user's presence, reed switches on cupboards and doors to measure open or close status, and float sensors in the bathroom to measure toilet being flushed or not. The data were annotated using two approaches: (1) keeping a diary in which the activities were logged by hand and (2) with the use of a blue tooth headset along with a speech recognition software. A total of three datasets were collected from the three different houses. Details about the datasets are shown in Table~\ref{datasets} where each column shows the details of the house with the information of the user living in it, the sensors placed in the house and the number of activity labels that were used. 
\begin{table}[tbh]
\centering
\caption{Details of the datasets.}
\begin{tabular}{cccc}
	\hline
\space& House A  &  House B  &  House C \\ %IP: maybe say participant instead of house?
	\hline
Age	&	$26$	&	$28$  &	   $57$ \\
Gender & Male   &   Male   &    Male\\
Setting& Apartment&Apartment & House \\
Rooms &   $3$		&	$2$		& $6$ \\
Duration &$25$days  & 	$14$days& $19$days \\
Sensors & $14$		&  $23$		& $21$ \\
Activities& $10$ & 	   $13$   &   $16$ \\
Annotation & Bluetooth& Diary & Bluetooth\\
	\hline
 
\end{tabular}
\label{datasets}
\end{table}

The data used in the experiments have different representation forms. The first form is raw sensor data, which are the data received directly from the sensor. The second form is last-fired sensor data which are the data received from the sensor that was fired last. The last firing sensor gives continuously 1 and changes to 0 when another sensor changes its state. For each house, we left one day out of the data to be used later for the testing phase and used the rest of the data for training. We repeated this for every day and for each house. Separate models are trained for each house since the number of sensors varies, and a different user resides in each house. Sensors are recorded at one-minute intervals for 24 hours, which totals in 1440 length input for each day. 

\subsection{Results}
\label{results}

The results presented in Table~\ref{rawtable} show the performance of the LSTM model on raw sensor data in comparison with the results of NB, HMM, HSMM and CRF \cite{kasteren2011human}. Table \ref{lastfired} shows the results of the LSTM model on last-fired sensor data again in comparison with the results of NB, HMM, HSMM and CRF. For the LSTM model, a time slice of ($70$) with hidden state size ($300$) are used. For the optimization of the network, Adam is used with a learning rate of $0.0004$~\cite{kingma2014adam} and Tensorflow was used to implement the LSTM network. The training took place on a Titan X GPU and the time required to train one day for one house is approximately 30 minutes, but training times differ amongst the houses. Since different houses have different days we calculated the average accuracy amongst all days. The training is performed using a single GPU but the trained models can be used for inference without losing performance when there is no GPU. 

%The experiments show the performance of the LSTM model on raw sensor and last-fired sensor data. To evaluate the results comparisons with the results of NB, HMM, HSMM and CRF are shown in Table \ref{rawtable} and Table \ref{lastfired}.% 

\begin{table}[tbh]
\centering
\setlength{\tabcolsep}{6pt}
\caption{Results of raw sensor data}
%\addtolength{\tabcolsep}{2pt}
\label{fig: tab1}
\begin{tabular}{cccc}
	\hline
Model & House A  &  House B  &  House C \\
	\hline
Naive Bayes	&	$77.1\pm20.8$	&	$80.4\pm18.0$  &	   $46.5\pm22.6$ \\
HMM& $59.1\pm28.7$  &  $63.2\pm24.7$   &    $26.5\pm22.7$\\
HSMM& $59.5\pm29.0$& $63.8\pm24.2$ & $31.2\pm24.6$ \\
CRF &   $89.8\pm8.5$		&	$78.0\pm25.9$		& $46.3\pm25.5$ \\
LSTM(Ours)& $\mathbf{89.8\pm8.2}$  & 	$\mathbf{85.7\pm14.3}$&  $\mathbf{64.22\pm21.9}$ \\
	\hline

\end{tabular}
\label{rawtable}
\end{table}

Table~\ref{fig: tab1} shows the results of different models on raw data from three different houses. The LSTM model has the best performance for all three data sets. In House B and House C, LSTM improves the best result significantly especially on House C where the improvement is approximately $40\%$.

\begin{table}[tbh]

\centering
\setlength{\tabcolsep}{6pt}
\caption{Results of last-fired sensor data}
\label{fig: tab2}
%\addtolength{\tabcolsep}{2pt}
\begin{tabular}{cccc}
	\hline
Model & House A  &  House B  &  House C \\
	\hline
Naive Bayes	&	$95.3\pm2.8$	&	$86.2\pm13.8$  &	   $87.0\pm12.2$ \\
HMM& $89.5\pm8.4$  &  $48.4\pm26.0$   &    $83.9\pm13.9$\\
HSMM& $91.0\pm7.2$& $67.1\pm24.8$ & $84.5\pm13.2$ \\
CRF &   $\mathbf{96.4\pm2.4}$		&	$\mathbf{89.2\pm13.9}$		& $\mathbf{89.7\pm8.4}$ \\
LSTM& $95.3\pm2.0$  & 	$88.5\pm12.6$&  $85.9\pm10.6$ \\
	\hline

\end{tabular}
\label{lastfired}
\end{table}

Table~\ref{fig: tab2} shows the results on last fired data from three different houses using the same models as in Table~\ref{fig: tab1}. The LSTM model did not improve the results in this section but it matched the best performance for two data sets with a slight drop in House C.  

\section{Discussion} 
\label{Discussion}
The results presented in this paper show that the deep learning based approaches for activity recognition from raw sensory inputs can lead to significant improvement in performance, increased accuracy, and better results. As shown in Section \ref{results} our LSTM based activity predictor matched or outperformed existing probabilistic models such as Naive Bayes, HMM, HSMM and CRF on raw input and in one case improved the best result by $40\%$. Predicting on raw input also reduces the human efforts required on data preprocessing and handcrafting features which can be very time consuming when it comes to an AAL (Ambient Assisted Living) environment. 

\section{Future Work} 
\label{Future Work}

Our future work will focus on reducing the variance on our predictions and early stopping criteria while training on different days. The LSTM model has different hyperparameters which affect the performance of the model significantly. Different optimization and hyperparameter search techniques could be investigated in the future. Since the LSTM model has proven to be superior on raw data it would be interesting to also apply other deep learning models. One problem is that deep learning badly captures model uncertainty. Bayesian models offer a framework to reason about model uncertainty. Recently, Yarin \& Ghahramani (2016) \cite{YarinGhahramani:2016:UncertaintyDeepLearning} developed a  theoretical framework casting dropout training in deep neural networks as approximate Bayesian inference in deep Gaussian processes. This mitigates the problem of representing uncertainty in deep learning without sacrificing either computational complexity or test accuracy. 

\section*{Acknowledgement}
This work has been funded by the European Union Horizon2020 MSCA ITN ACROSSING project (GA no. 616757). The authors would like to thank the members of the project's consortium for their valuable inputs.

\bibliographystyle{splncs}
\makeatletter
\renewcommand\@biblabel[1]{#1. }
\makeatother
%----------------------------------------------------------
% Use the following option to include external BibTeX files:
\bibliography{references}

\end{document}